\def\year{2017}\relax
\documentclass[letterpaper]{article}
\usepackage{aaai17}
\usepackage{times}
\usepackage{helvet}
\usepackage{courier}
\begin{document}

\title{Blue Sky Ideas in Artificial Intelligence Education\\from the EAAI 2017 New and Future AI Educator Program}
\author{
Eric Eaton\\
Univ.~of Pennsylvania\\
eeaton@cis.upenn.edu\\
\And
Sven Koenig\\
U.~Southern California\\
skoenig@usc.edu\\
\And
Claudia Schulz\\
Imperial College London\\
claudia.schulz@imperial.ac.uk\\
\And
Francesco Maurelli\\
Jacobs University~Bremen\\
f.maurelli@ieee.org\\
\AND
John Lee\\
Antioch Univ.\\
John@AssistiveIntelligence.com\\
\And 
Joshua Eckroth\\
Stetson Univ.\\
jeckroth@stetson.edu\\
\And
Mark Crowley\\
Univ.~Waterloo\\
mcrowley@uwaterloo.ca\\
\And
Richard G. Freedman\\
U.~Massachusetts  Amherst\\
freedman@cs.umass.edu\\
\AND
Rogelio E. Cardona-Rivera\\
North Carolina State U.\\
recardon@ncsu.edu\\
\And
Tiago Machado\\
New York Univ.\\
tiago.machado@nyu.edu
\And
Tom Williams\\
Tufts Univ.\\
williams@cs.tufts.edu\\
}
\maketitle

\begin{abstract}
The 7th Symposium on Educational Advances in Artificial Intelligence (EAAI'17, co-chaired by Sven Koenig and Eric Eaton) launched the EAAI New and Future AI Educator Program to support the training of early-career university faculty, secondary school faculty, and future educators (PhD candidates or postdocs who intend a career in academia).  As part of the program, awardees were asked to address one of the following ``blue sky'' questions: 
\begin{enumerate}
    \item How could/should Artificial Intelligence (AI) courses incorporate ethics into the curriculum?
    \item How could we teach AI topics at an early undergraduate or a secondary school level?
    \item AI has the potential for broad impact to numerous disciplines. How could we make AI education more interdisciplinary, specifically to benefit non-engineering fields? 
\end{enumerate}
This paper is a collection of their responses, intended to help motivate discussion around these issues in AI education.
\end{abstract}

\section{Bridging Across Disciplines}
\begin{center}
{\bf Claudia Schulz} (Imperial College London)
\end{center}

The application of AI methods to problems such as legal decision making, language translation, or gene analysis often requires the cooperation of AI experts and subject specialists, e.g., lawyers, translators, or biologists. Their ability to communicate on a common ground is a crucial factor determining the success of the project. It is thus beneficial if both parties have a basic understanding of the subject as well as of AI methods, even before the start of a project.

Universities provide a unique opportunity to both teach students becoming AI experts some subject knowledge (e.g., biology or law) and ensure that students in non-computing subjects have a basic understanding of AI techniques. A na\"{i}ve approach for achieving such interdisciplinary learning is that AI students take some first-year subject courses, and subject students some introductory AI courses. Even though this approach is easy to implement, it may not achieve the intended interdisciplinary learning benefits since the courses are not tailored towards students of a different discipline (even first-year courses often provide a detailed introduction to a specific topic instead of surveying a whole field). 

We here discuss two approaches based on \emph{peer-learning}, which provide a more beneficial interdisciplinary learning environment. They share the idea that AI and subject students learn together by teaching each other. 

In the \emph{seminar-style approach}, AI students give seminars to subject students (and vice versa). These seminars may, for example, provide an overview of AI techniques or review applications of AI methods in subject areas. This approach does not only benefit the attending subject students, who acquire knowledge tailored particularly to them, but also provides valuable experience to the AI student giving the seminar in explaining AI topics to the lay audience. There is clearly a lot of variability concerning the exact setup of these seminars: they can be given by a single PhD student or by a group of undergraduates, and the attendees' background can be a mixture of subjects or a single subject (in which case the seminar will cover topics and examples related to this particular subject).

In contrast to the seminar-style approach, where the speaker teaches the audience, the \emph{project-based approach} promises mutual teaching and learning, both in terms of knowledge and skills. In this setting, an AI student and a subject student work together on a project trying to solve a problem in the subject student's area by applying AI techniques. At the start, the subject student explains subject-specific background to the AI student, whereas the AI student teaches the subject student about possible AI techniques to be used, thus creating a mutual teaching and learning environment. During the project, students will also acquire the invaluable skills of working in an interdisciplinary team. Again, there are different setups for such projects: The problem(s) to be solved can be given by faculty or be the students' own ideas, and the project can be part of a course or an extra-curricular ``ideas/start-up lab''.

\section{Student-Centric Discovery}
\begin{center}
{\bf Francesco Maurelli} (Jacobs University Bremen)
\end{center}

Most approaches in university teaching are based on frontal lectures, sometimes with specific lab activities and specific homeworks. The course is divided in specific modules which are explained sequentially. 

I would be interested in analysing the feasibility (and try that with a real course) of a more student-centric approach, inspired by the pedagogical Montessori method (Montessori and George 1964). Although the main focus of the method has always been on children, some of those elements have been incorporated with success in secondary-school and early-undergraduate levels.

Working with an equipped lab is fundamental for this approach. Then I would imagine that each student (or maybe each group of students) could freely decide the direction of the course, based on discovery and on what they are interested in. I have recently started a cooperation with Prof. Federico Gobbo at the University of Amsterdam, to analyse the portability of some key elements of the Montessori method into AI education of young adults (so, the target group of this call).

From one side, I am interested to see how the Montessori method applied at a later age group than usual could help the students in their personal and professional development. I strongly believe that independence and the ability of thinking, reasoning, and making informed choices are key elements of the lives of active and engaged human beings, part of the society. A teaching approach which values independent thinking seems therefore a very interesting and potentially fruitful approach, albeit maybe difficult at times.

From the other side, looking for new engaging methods of teaching AI and robotics might result in students approaching the subject with curiosity and willingness, not just because it is in the study plan. This in turn might result in more people engaged in AI and Robotics, and in more passion towards the subject. It might be perceived not just as one of many lectures, but a feel of ``ownership'' might push for a deeper understanding of specific subjects rather than usual frontal lectures.

Challenges in this approach would be ensuring that each student (or each group of students) progress and explore the subject within some boundaries. Also, evaluation is a very delicate subject. In the original Montessori approach there is no grading for children, but it is something usually necessary in undergraduate courses. Establishing a fair grading system is something necessary, though it might be hard to compare different approaches and different paths that each student would undertake.\\[-0.75em]

{\footnotesize 
\noindent {\bf References}

\noindent Montessori, M.; and George, A.E. 1964. {\em The Montessori method.} New York: Schocken Books.\vspace{0.25em}

\noindent Dohrmann, K.R.; Nishida, T.K.; Gartner, A.; Lipsky, D.K.; and Grimm, K.J. 2007. High school outcomes for students in a public Montessori program. {\em Journal of Research in Childhood Education} 22(2): 205--217.
}

\section{AI in K-12 Education}
\begin{center}
{\bf John Lee} (Antioch University)
\end{center}

I believe that we are fundamentally overlooking the development of important concepts in K-12 education.  Both in non-arithmetic skill building as well as unique perspectives on ourselves.

One of the major over-arching themes in K-12 education is the development of humanity.  Early cultures tell us about our society and the origins government.  Zoology tells us not only about the animal kingdom, but also what it means to be human.  Mathematics teaches us about fundamental truths and beauty.  Astronomy teaches us our place in the cosmos and inspires us to reach beyond our own limitations.

AI can also teach us about what it means to be human.  It can teach us what humanity looks like when taken to different extremes and thus develop within ourselves a deeper understanding of each other and our differences.  It can easily demonstrate the truth and beauty of mathematics and how it can be used to develop models of knowledge and behavior.  Each one of these models can then provide us with a unique perspective into our own cognition, psychology and the perspective of our existence.

A solid foundation in mathematics will start with movement, which will flow from real object manipulation to imagination to abstract cognition.  This is introduced with early arithmetic.  However, there is no similar early introduction of non-arithmetic cognition such as logic, search, iteration (folding), etc... that are vital for all kinds of engineering and programming.  Such professions are shown to be deeply imaginative from mentally stepping through a program's execution to predicting the voltage levels across a circuit diagram.  Early introduction of agent-based models through games and puzzles could provide this foundation as well as begin to introduce concepts for later exploration such as search, string-replacement-iteration, planning, machine learning, etc.

What this solid foundation of movement and imagination provides is a deeper understanding of and greater passion for mathematics.  By the time we get to techniques such as multiple-column multiplication or long division we are beginning to learn procedures.  This is what will make or break the love of mathematics.  Those that truly learn what is behind the procedure and can see it in their imagination will do well, those that learn to blindly follow the procedure will not.

By the time we get to the upper grades, so much of engineering and sciences are taught procedurally.  It is a crucial time to emphasize the importance of true-understanding, but class sizes, time constraints and material creep will make this difficult.  How much easier would it be if there is a number of early grades experiences that begin to magically resonate with what is being taught.

I hope to explore the earliest introduction of core AI concepts in a concrete way to develop technical imagination skills, get us to think about how we think and finding new confidences in ourselves as we explore what it means to be human.

\section{The Role of Ethics in AI Education}
\begin{center}
{\bf Joshua Eckroth} (Stetson University)
\end{center}

In CS education, and AI education in particular, ethics is too often treated as a side concern, addressed in isolation from more typical topics. We view ethics as a cross-cutting concern that helps inform AI students, researchers, and practitioners how to be good scientists and engineers. Here, we examine five topics that are typically included in an AI curriculum and their respective ethical dimensions.

{\em (1) Search and planning}:~~AI systems that are deployed into real-world settings will be expected to perform accurately and reliably. Consider a search procedure, marketed as ``Astar,'' that does not always find an optimal path due to a non-admissible heuristic. Or consider a planning system that does not account for the ``frame problem,'' makes a wrong assumption about the state of the world, and fails to observe before acting. These examples illustrate unquantified risk resulting from inappropriate algorithmic decisions.

{\em (2) Knowledge representation (KR) and reasoning}:~~A KR schema is a surrogate for real-world entities (Davis, et al. 1993), and rarely attempts to model all of their complexities. For example, discretizing the range of human relationships into friends, married, or ``it's complicated'' introduces ethical questions about whether and what kind of inferences can be accurately drawn. Yet, high fidelity representations and inferential expediency remain in constant tension.

{\em (3) Probabilistic reasoning}:~~While probabilistic knowledge helps avoid making strict claims when knowledge is insufficient, probabilistic reasoning rarely yields certain inferences. Instead, some kind of decision theory must be consulted, which brings ethical questions about estimating risk and utility.

{\em (4) Machine learning (ML)}:~~Learned models can be difficult to trust due to their complexity. In this sense, interpretable models like decision trees are less risky than less-interpretable models like neural networks. In either case, trust can be enhanced with holdout and cross validation techniques. ML is more than ``picking the technique that gives highest accuracy.'' We should know that the technique is best suited to the task at hand, and be able to justify that decision.

{\em (5) Robotics}:~~Once equipped with actuators, robots enter the ethical dimension. Failing to send a ``stop motor'' command due to software flaws may result in disastrous consequences. Machine/human control handoff (Klein et al. 2004), sometimes realized as a big red button, is a moment of vulnerability that can be mitigated with better status reporting and situation awareness. These issues go beyond typical robot building challenges.

We have shown that ethics should be addressed throughout the AI curriculum. The need for ethics arises from the need to be sure we are building systems that are appropriate for real-world situations and usable by people who depend on their accurate and reliable functioning.\\[-0.75em]

{\footnotesize 
\noindent {\bf References}

\noindent Davis, R.; Shrobe, H.; and Szolovits, P. 1993. What is a knowledge representation? {\em AI Magazine} 14(1): 17--33.\vspace{0.25em}

\noindent Klein, G.; et al. 2004. Ten challenges for making automation a team player in joint human-agent activity. {\em IEEE Int Sys} 19(6): 91--95.
}

\section{AI Education through Real-World Problems}
\begin{center}
{\bf Mark Crowley} (University of Waterloo)
\end{center}

It is increasingly essential that practitioners of AI and ML focus on building verifiable tools with solution bounds or guaranteed optimality. One of my aims for AI/ML students is to give them the skills to build algorithms and analytical tools for providing verifiable guarantees and quality bounds on classification, prediction, and optimization problems. 

The usual approach in a maturing field such as AI/ML would be to establish engineering standards for tools and methodologies that provide verifiable quality bounds and guarantees. Yet, the development of relevant tools are still an emerging research pursuit. Witness the extensive interest in the probability bounded results of Bayesian Optimization, the expanding application of PAC learning algorithms, or the wide usage of Gaussian processes to represent uncertainty and guide efficient sampling. 

There is a growing application of AI/ML algorithms to safety critical domains such as automated driving, medical decision making and analysis and financial management. Also critical is the growth of computational sustainability: application of AI/ML methods to natural resource domains, wildlife management, energy management, socioeconomic planning and climate modelling. These domains all involve huge societal investments and impacts. Planning is often over a long horizon so what are acceptable risks and uncertainties in short term problems can expand over time into huge errors which undermine results.

Teaching students about these problems and the tools to address them will have an immediate impact on the world. In AI education we need to develop a new nucleus of an engineering discipline for AI/ML that provides students the framework to navigate the ever-expanding set of computational tools for solving complex problems.  

This is an education ethics issue as well. If we are turning out students with the answers to the world's problems, they need to know how to justify those answers in a rigorous way. This is often not the primary focus of AI/ML research or education. Many students who study AI/ML will continue straight on to industry rather than further research, so they will need to know the best algorithms to apply to different classification, prediction, optimization problems. However, to be AI engineers will in a way will require students to know more theory than a fully applied program which teaches use of existing methods. They need to know enough about the underlying probabilistic model, the sample complexity and the relationship of prior, latent, and observed variables in order to understand how reliable the results of their models are.  Students also need a strong grounding in classical as well as Bayesian statistics so that they can make the right methodological choices for the given situation and do more than simply showing a histogram or ROC curve for their problem to justify their performance.

So, I feel the future of AI/ML education, especially at the undergraduate and master's level, is increasingly going to be focused on making AI into a true engineering discipline where requirements, guarantees, and design are as critical as reducing raw error rates of a classifier.

\section{Making AI Concepts More Accessible}
\begin{center}
{\bf Richard G. Freedman} (Univ.~of Massachusetts Amherst)
\end{center}

While it may be unreasonable to expect early undergraduate and secondary school students to code AI algorithms, it is possible for them to visualize and experience these algorithms firsthand.  Developing an understanding of AI through these perspectives may even facilitate abstract thinking and problem solving when learning computer science and programming later.  Although taught later in the CS curriculum after students are comfortable with computational thinking, many topics in AI can be explained conceptually using only high school mathematics.  However, the manner in which these concepts are taught needs to be less traditional.

Based on the average student's present-day lifestyle involving personal mobile devices and almost limitless access to media, most students are used to constantly interacting with others and/or engaging in entertainment.  This nearly contradicts the traditional lecture style for presenting material impersonally at the front of the room using chalkboards or slides.  Instead, students today are accustomed to short spurts of watching and then lots of time doing, which goes hand-in-hand with some elements of team-based learning.  In particular, an instructor should only briefly introduce a topic and related activity.  Then, the students may explore the activity in groups in order to experience the concept on their own, interacting with each other to understand what happens.  For example, A* search can be performed with a map and deck of cards; each card covers a city and students write the ruler distance (Euclidean heuristic) on each card as it is added to the frontier.  The visited cities' cards are stacked in a deck to visualize the visited sequence.

By focusing on the algorithms' processes rather than the specific implementation, younger students without computational experience, higher-level mathematics, and programming skills can participate.  The early focus of AI was to emulate human intelligence, and these students can relate to that by wondering, "how would I solve this problem?"  These are questions they can discuss with each other and the instructor while performing the activities.  In particular, the instructor can now make her time with students more personal by visiting groups to discuss and give tips based on their progress.  Groups can also interact with each other afterwards to compare results. 

Just as important as the interaction in the classroom, time outside of class can be vital to learning.  Besides homework assignments that review concepts, students spend time on the internet watching videos and listening to music.  Educational content can be provided in such entertaining forms.  Alongside the classic television series {\em Bill Nye the Science Guy}, on-line streaming services such as YouTube have channels devoted to fun, short videos teaching mathematics (Vihart) and science (Veritasium, VSauce).  While such a channel does not seem to exist for AI outside of Michael Littman's music videos, it is possible to present real-world examples and perform the activities above to create one. Then younger students are exposed to AI topics at any time in formats that they are more ready to digest, using high school-level knowledge without focusing on the code.

\section{Rethinking the AI Ethics Education Context}
\begin{center}
{\bf Rogelio E. Cardona-Rivera} (North Carolina State U.)
\end{center}

Ethics, the moral principles that govern a person's or group's behavior, cannot be incorporated into a curriculum around AI without a systematic revision of the surrounding context within which AI takes place. We must go beyond just talking about ethics in the classroom; we need to put ethics into practice. I offer three recommendations for doing so, drawn from how ethics are treated within engineering and the social sciences. 

Firstly, the Association for the Advancement of AI (AAAI) should institute an association-wide code of ethics. This recommendation is inspired by ethics codes in engineering, which include concern for the public good as a constituent part.  For instance, the code of ethics of the National Society of Professional Engineers (2007) contains seven fundamental canons, the first of which is: ``Engineers, in the fulfillment of their professional duties, shall hold paramount the safety, health and welfare of the public.'' An association-wide code of ethics would formally recognize our impact in and the responsibility that we owe to our society.

Secondly, research funding applications that deal with AI should be required to assess risks to society. This recommendation is inspired by similar requirements by Institutional Review Boards (IRB) within the social sciences (e.g., U.S. Dept.~of Health and Human Services 2009). Whenever researchers conduct studies that deal with human participants, they are asked by an IRB to assess sources of potential risk; AI research applications should do the same. Importantly, these risk assessments should consider threats beyond immediate physical harm; e.g., the development of new analytical tools for understanding large amounts of data may inadvertently make it easier to reconstruct personally identifiable information, which constitutes a threat to anonymity, and which may disadvantage vulnerable populations.

Thirdly, students in AI project-based courses should be required, as part of the class' deliverables, to submit documents that assess the impact to society (in the context of the proposed AAAI code of ethics, and which should include an IRB-like risk assessment). Ideally, AAAI would serve as a facilitator of this kind of assessment, by providing a library of case studies and expert testimonies that can guide students in examining the broader implications of their work.

Incorporating ethics into a curriculum is more than a one-shot affair. It requires a systematic revision of the surrounding context within which AI exists, in terms of how we talk about it (first recommendation), how we fund efforts in it (second recommendation), and how it is put into practice (third recommendation). By leveraging existing models on ethics from engineering and the social sciences, we will be better equipped to offer concrete recommendations to ensure that ethics aren't an afterthought, but are integral to the development of AI.\\[-0.75em]

{\footnotesize
\noindent {\bf References}

\noindent National Society of Professional Engineers. 2007. {\em Code of ethics for engineers}. Technical Report 1102.\vspace{0.25em}

\noindent U.S.~Department of Health and Human Services. 2009. 45 CFR 46.111: {\em Criteria for IRB approval of research}. Technical report.\vspace{0.25em}
}

\section{Lifelong Kindergarten for AI}
\begin{center}
{\bf Tiago Machado} (New York University)
\end{center}

To meet the expectations of young generations (who are highly exposed
to games and other virtual interactions) 
regarding an introductory AI course, our purpose
is to design a course based on the principles of the Lifelong Kindergarten (LK) (Resnick, M. 2007) and the Zone of Proximal Development (ZPD) as fields for dialogue (Meira and Lerman. 2001). From the former, we follow the principles of imagine, create, share and reflect. From the latter, we follow the idea of using it as a way to improve class communication with and among students.

We plan to create a 12-week course (2-hour class) in a
level suitable for secondary school students with previous
experience in programming languages. As video games
are an attractive media to our audience, the course
will use the General Video Game Framework (GVG-AI) (Perez-Liebana et al. 2016), which allows developers to implement
algorithms to play famous arcade games. The
course will have three stages: 1) Introduction to the
GVG-AI, 2) Search Algorithms and 3) Supervised
Learning Algorithms.

The first stage (Introduction to GVG-AI) explains
how to work with the framework. It guides the students
through a set of simple examples, followed by
simple assignments, like creating an agent that plays
the games by choosing random actions. The second and the third stages present the same
structure: in the first week, the instructor explains the
algorithms. Afterward, the students will have three
weeks to implement the algorithm assigned to their
group plus a class presentation. During these weeks
the course will work in a blending class format. The
students will have total access to videos, books, software, 
and the instructor to study and learn how to implement
the algorithms in the GVG-AI framework.

It will be required that the presentation
should not be a traditional one (i.e.,  students
presenting slides and speaking about what they did). Fun
and play with the content should be encouraged. As
well as taking extra care about actually teaching to others
how they can obtain the same results.

This way the students will be more active by imagining
and creating their solutions. 
During the presentations,
we will exercise more the share and reflect principles of
the LK. The students will be encouraged to ask questions
and share (all the resources they used to learn and
implement, including the resulting code) their solutions
with the class.

Throughout the course, instructors should be aware of students' progress. They should create social network
or email channels to connect with students both 
in- and outside class. In this way we exercise, in
both physical and virtual situations, the ZPD function
of an intersubjective space via activities in
which participants teach and learn from
each other.

\vspace{0.25em}

{\footnotesize 
\noindent {\bf References}

\noindent Meira, L.; and Lerman, S. 2001. {\em The zone of proximal development as a symbolic space.} South Bank University.\vspace{0.25em}

\noindent Perez-Liebana, D.; et al. 2016. The 2014 general video game playing competition. {\em IEEE Trans.~Comp.~Intel.~and AI in Games} 8(3).\vspace{0.25em}

\noindent Resnick, M. 2007. All I really need to know (about creative thinking)
I learned (by studying how children learn) in kindergarten. {\em ACM SIGCHI Conf.~on Creativity \& Cognition} (C\&C '07).

}

\section{Training Students in AI Ethics}
\begin{center}
{\bf Tom Williams} (Tufts University)
\end{center}

After completing a course in AI, it is generally assumed that a student will be able to (1) characterize the task environment of a new problem, including the performance measure which should be optimized in that environment, and (2) identify the design tradeoffs between different algorithms for solving that problem. Unfortunately, students  are rarely taught to consider the ethical facets of task environments that should be taken into account when deciding on performance measures and considering design tradeoffs, leading to blind spots for ethical failures in algorithm design.

In order to remove this blind spot, I believe that educators should strive to achieve the following learning objective: Students should be able to identify   circumstances in which a tradeoff must be made with respect to task performance and ethical performance (especially with an eye towards verifiability), and be able to argue why a particular choice of algorithm strikes an appropriate balance between task performance and ethical performance.

One way to fulfill this objective could be to train students to evaluate proposed AI solutions by asking the following:

{\em Consequentialism:}~~
     (1) Is it possible that a decision made within this problem domain
        could harm another agent?
     (2) If so, can you guarantee that the proposed approach will find the
        solution that does the least harm (or harm below
        some justifiable threshold)?
     (3) If the answer to 2 is no, is there any other
        known AI solution for which the answer is yes?
     (4) If the answer to 3 is yes, what is the justification for the
        use of the proposed algorithm? If the answer to 3 is no,
        what is the justification for solving this problem computationally?

{\em Deontology:}~~
     (1) Is it possible that a decision made within this problem domain
        could violate a legal statute or moral norm?
     (2) If so, can you guarantee that the proposed approach will find
        a solution that results in the fewest rule violations (or violations below some justifiable threshold)?
     \mbox{(3--4)}~Same as above.
     
{\em Virtue Ethics:}~~
     (1) Is it possible that a decision made within this problem domain
        could be legal, and avoid explicit harm, yet fail to align
        with human virtues?
     (2) If so, can you guarantee that the proposed approach will find
        a solution that results in optimally virtuous behavior (or achieves a level of virtue that is above some
        justifiable threshold)?
     \mbox{(3--4)}~Same as above.

The purpose of using this framework is to force students to ``think like an ethicist'' when designing or choosing between AI solutions: even though ethical concerns often present moral dilemmas to which there is no single obviously correct solution, students should get used to analyzing proposed solutions in order to identify possible ethical problems, identify what types of AI solutions make it difficult to verify or quantify ethical performance, and convincingly argue for or against a potential solution on ethical grounds.

\section*{Acknowledgements}

The EAAI'17 New and Future AI Educator Program is partly supported by NSF Grant \#1650295 and funding from the Artificial Intelligence Journal.

\end{document}